\documentclass[journal]{IEEEtran}

\usepackage{cite}
\ifCLASSINFOpdf
  \usepackage[pdftex]{graphicx}
\else
  \usepackage[dvips]{graphicx}
\fi
\usepackage{amsmath}
\usepackage{color}
\usepackage{amssymb}
\usepackage{algorithm,algorithmic}
\usepackage{array}
\ifCLASSOPTIONcompsoc
  \usepackage[caption=false,font=normalsize,labelfont=sf,textfont=sf]{subfig}
\else
  \usepackage[caption=false,font=footnotesize]{subfig}
\fi
\usepackage{wasysym}
\usepackage{multirow}
\usepackage{enumerate}
\usepackage{comment}
\usepackage{verbatim}
\usepackage{soul}

\newtheorem{assumption}{A}

\newtheorem{observation}{Observation}

\begin{document}

\title{On-line Building Energy Optimization using \\Deep Reinforcement Learning}

\author{Elena~Mocanu,
         Decebal~Constantin~Mocanu,
         Phuong~H.~Nguyen,
         Antonio~Liotta,
         Michael~E.~Webber,
        ~Madeleine~Gibescu,
        ~J.G.~Slootweg
     
\thanks{E. Mocanu, D.C.Mocanu, P.H.Nguyen, A.Liotta, M.Gibescu and J.G. Slootweg are with the Department of Electrical Engineering, Eindhoven University of Technology, Eindhoven, 5600 MB, The Nedherlands.(e-mail:\{e.mocanu; d.c.mocanu; p.nguyen.hong; a.liotta; m.gibescu; j.g.slootweg\}@tue.nl)}
\thanks{Michael E. Webber is with the Department of Mechanical Engineering, The University of Texas at Austin, Austin, TX 78712-1591, USA. (e-mail: webber@mail.utexas.edu)}
}

\markboth{This manuscript is a pre-print version!}
{Shell \MakeLowercase{\textit{et al.}}: Bare Demo of IEEEtran.cls for IEEE Journals}

\maketitle

\begin{abstract}
Unprecedented high volumes of data are becoming available with the growth of the advanced metering infrastructure. These are expected to benefit planning and operation of the future power system, and to help the customers transition from a passive to an active role. In this paper, we explore for the first time in the smart grid context the benefits of using Deep Reinforcement Learning, a hybrid type of methods that combines Reinforcement Learning with Deep Learning, to perform on-line optimization of schedules for building energy management systems. The learning procedure was explored using two methods, Deep Q-learning and Deep Policy Gradient, both of them being extended to perform multiple actions simultaneously. The proposed approach was validated on the large-scale Pecan Street Inc. database. This highly-dimensional database includes information about photovoltaic power generation, electric vehicles as well as buildings appliances. Moreover, these on-line energy scheduling strategies could be used to provide real-time feedback to consumers to encourage more efficient use of electricity.

\end{abstract}

\begin{IEEEkeywords}
Deep Reinforcement Learning, Demand Response, Deep Neural Networks, Smart Grid, Strategic Optimization
\end{IEEEkeywords}

\IEEEpeerreviewmaketitle

\section{Introduction}
\IEEEPARstart{T}{here} is an energy transition underway since the start of the millennium, comprised primarily of a push towards replacing large, fossil-fuel plants with renewable and distributed generation. It results in increased uncertainty and complexity in both the business transactions and in the physical flows of electricity in the smart grid. Because the built environment is the largest user of electricity, a deeper look at building energy consumption holds a promise for improving energy efficiency and sustainability. Understanding such individual consumption behavior based on the knowledge transfer from the fusion of extensive data collected from the Advanced Metering Infrastructure (AMI) is an essential step to optimize building energy consumption and consequently the effects of  its use.

This work is motivated by the hypothesis that an optimal resource allocation of end-user patterns based on daily smart electrical device profiles could be used to smoothly reconcile differences in future energy consumption patterns and the supply of variable sources such as wind and solar~\cite{Alam,Barbato_2014,tps2016crisdent}. It is expected that a cost minimization problem could be solved to activate real-time price responsive behavior~\cite{tse2010bun}. A wide-range of methods have been proposed to solve the building energy and cost optimization problems, including linear and dynamic programing, heuristic methods such as Particle Swarm Optimization (PSO), game theory, fuzzy methods and so on~\cite{Barbato_2014,Nikos2016, Alam,tps2016crisdent,tse2010bun,Vardakas2015,luisSMARTGREEN}. Therein, both centralized and decentralized solutions exist, but they fail to consider on-line solutions for large-scale, real databases~\cite{Vardakas2015}. More concretely, any time when an optimization is needed, these methods have to compute completely or partially all the possible solutions and to choose the best one. This procedure is time consuming. In the big data era, more and more machine learning methods appear to be suitable to overcome this limitation by automatically extracting, controlling and optimizing the electrical patterns. This can be done by performing successive transformation of the historical data to learn powerful machine learning models to cope with the high uncertainty of the electrical patterns. Then, these models will be capable of generalization and they could be exploited in an on-line manner (i.e. few milliseconds) to minimize the cost or the energy consumption in newly encountered situations. Among all these machine learning models, the ones belonging to the Reinforcement Learning (RL) area are the most suitable for the cost minimization problem, as they are capable to learn an optimal behavior, while the global optimum is not known.

Thus, in the remaining of this paper we focus on RL methods, such as Q-learning~\cite{Watkins}, and their latest developments. The building environment is modeled using a Markov Decision Process~\cite{Sutton:1998} and it can be used to find the best long-term strategies. Prior studies showed that RL methods are able to solve stochastic optimal control problems~\cite{damian}  in the power system area as well as an energy consumption scheduling problem \cite{5622078} with dynamic pricing~\cite{tse2016RL}.  A batch reinforcement learning method was  introduced in \cite{7038106} to schedule a cluster of domestic electric water heaters, and further on applied for smart home energy management~\cite{ijcai2015}. Owing to the curse of dimensionality, these methods fail for large-scale problems. More recently, there has been a revival of interest in combining deep learning with reinforcement learning. Lately, in 2015, an application of Q-learning to deep learning has been successful at playing Atari2600 games at expert human levels~\cite{mnih15}. In 2016, another one has defeated for the first time in history the world champion at the game of Go. Complementary with our work, Francois-Lavet et al. has proposed the use of Deep Q-learning for storage scheduling in microgrids~\cite{franccois2016deep}. The above methods represent the starting point of a new research area, known as Deep Reinforcement Learning (DRL), which has evolved through the intersection of reinforcement learning and neural networks. At the same time, in our previous work, we showed that Reinforcement Learning using Deep Belief Networks for continuous states estimation can successfully perform unsupervised energy prediction~\cite{Mocanu2016646}.

\textit{Our contribution:}
In this paper, inspired by the above research developments, we propose for the first time the use of the Deep Policy Gradient method, as part of Deep Reinforcement Learning algorithms, in the large-scale physical context of smart grid - smart building, as follows. 
\begin{itemize}
\item We propose a new way to adapt DRL algorithms to the smart grid context, with the aim of conceiving a fast algorithm to learn the electrical patterns and to optimize on-line either the building energy consumption or the cost.
\item We investigate two DRL algorithms, namely Deep Q-learning (DQN)~\cite{mnih15} and Deep Policy Gradient (DPG). 
\item DPG in its current form is capable to take just one action at a specific time. As in the building context multiple actions have to be taken at the same moment, we propose a novel method to enhance DQN with the capability of handling multiple actions simultaneously.
\end{itemize}
We evaluate our proposed methods on the PecanStreet database at both the building and aggregated level. In the end, we prove that our proposed methods are able to efficiently cope with the inherent uncertainty and variability in the generation of renewable energy, as well as in the peoples' behavior related with their use of electricity (i.e. charging of electric vehicles). Specifically, we show that the enhanced DPG is more appropriate to solve peak reductions and cost minimization problems than DQN.

The remaining of this paper is organized as follows. Section II describes the problem formulation and Section III we introduce the background and preliminary concepts. Section IV describes our proposed method followed by implementation details in Section V. Results and discussions are provided in Section VI. Finally, we conclude with some directions for future research.

\section{Problem formulation}
In this context, we aim to reduce load peaks as well as to minimize the cost of energy.
 Let $\mathcal{B}$ denote the set of buildings, such that $B_{i} \in \mathcal{B}, \forall i \in \mathbb{N}$ representing the index of the building analyzed. The total building energy consumption $E_i$ is a sum over all power generation $P^{+}$ and consumption in a specific interval of time $\Delta t$. Therein, based on the shifting capabilities of appliances present in a building we differentiate between flexible power $P^{-}_{d}$, e.g. electric devices $d\in\{1,..,m_i\}$, and fixed consumption $P^{-}$. 
\paragraph{Cost minimization problem}
In this paper, we assume two price components over the space of $\mathcal{B}$, such that $\lambda_{t}^{-}$ is the price value set by the utility company for the time-slot $t$ and $\lambda_{t}^{+}$ represents the price value at which the utility company buys energy from end-users at time-slot $t$. Therefore, the optimal cost associated with customer $i$ at time $t$ for an optimization time horizon $T$ can be calculated as 
\begin{align}\label{eq:cost}
& \underset{}{\text{min}}
& &\sum_{t=1}^{T}(\lambda_{t}^{+}\sum_{i=1}^{n}P^{+}_{i,t}-\lambda_{t}^{-}\sum_{i=1}^{n}(P^{-}_{i,t}+\sum_{d=1}^{m_i}a_{i,d,t}P^{-}_{i,d,t}))\\
& \text{s.t.}
& & \sum_{t=1}^{T}P^{-}_{i} \Delta t =E_{i}, \; \forall i \in \mathbb{N}, \forall t \in \mathbb{N}, \\
& && \sum_{t=1}^{T}P^{-}_{d} \Delta t =E_{d}, \; \forall d \in \mathbb{N}, \forall t \in \mathbb{N}, \\
& && a_{i,d,t}=\{1,0\}, \forall a\in\mathcal{A},\forall i \in \mathbb{N}, \forall d \in \mathbb{N}, \forall t \in \mathbb{N},\\
&&&P^{+}_{i,t},P^{-}_{i,t}, P^{-}_{i,d,t} \geq 0, \forall t=[1:T] \in \mathbb{N},\\ \label{eq:const}
&&& \lambda_{t}^{+}, \lambda_{t}^{-} \geq 0, \forall t=[1:T] \in \mathbb{N}.
\end{align}
where $a_{i,d,t}=1$ if the electrical device is $on$ at that specific moment in time, and $0$ otherwise. Please note that, in our proposed method, computing $a_{i,d,t}$ is equivalent with the estimation of the actions (see Fig.1).

\paragraph{Peak reduction problem}
In the special case of constant price, for electricity generation and consumption, with $\lambda_{t}^{+}=\lambda_{t}^{-}$, the cost minimization problem becomes a peak reduction problem, defined as
\begin{align}\label{eq:12}
&\underset{}{\text{min}}
&&\sum_{t=1}^{T}\big(\sum_{i=1}^{n}P^{+}_{i,t} -\sum_{i=1}^{n}(P^{-}_{i,t} +\sum_{d=1}^{m_i}a_{i,d,t}P^{-}_{i,d,t})\big)
\end{align}
Consequently, the constraints following Eq.~\ref{eq:cost} will remain valid for both problems. However, based on the differences between different types of electrical devices the full range of constraints is larger as explained in the next sections.

\paragraph{Electrical device constrains}
We are assuming three types of consumption profiles. Firstly, we consider the $\textit{time-scaling load}$. In respect to this we confine our analysis to the air conditioning load $(d_{AC})$, as a representative part of a larger set of electrical devices in every building which could be switched on-off for a limited number of times during an optimization horizon, e.g. lights, television, refrigerator. Prior studies show that short-term air conditioning curtailments have a negligible effect on end-user comfort~\cite{aircond}.
Secondly, we include the $\textit{time-shifting load}$, also called deferrable load, that must consume a minimum amount of power over a given time interval. Therein, we model the dishwasher $(d_{DW})$ as an uninterruptible load, which requires a number of consecutive time steps. 
Finally the electric vehicle$(d_{EV})$ was considered as both a \textit{time scaling and shifting load}. A more rigorous formulation of the building electrical components and their associated constrains could be found in~\cite{OPT-002}. In our case, a complementary probabilistic perspective over the time dependent devices constrains $a_{d,t}$ give us the following assumptions:
\begin{assumption}\label{as:1}
For all $d$, with $P^{-}_{d}$ time-scaling loads, there $\exists\delta_{d}\in \mathbb{R_+}$ constants over the optimization horizon such that
\end{assumption}
\begin{equation}\label{eq:ass1}
\begin{cases}
 \sum_t  P^{-}_{d}\leq\delta_{d} &\text{if $p(P^{-}_{d}=0|t)\in (0,1]$}\\
 \sum_t  P^{-}_{d}=\delta_{d} &\text{if $p(P^{-}_{d}=0|t)=0$}
\end{cases}
\end{equation}
where $p(P^{-}_{d}=0|t)$ is the probability of the electrical device $d$ to be active at any moment in time $t$, for all $t=[1:T] \in \mathbb{N}$.
\begin{assumption}\label{as:2}
For all $d$, with $P^{-}_{d}$ time-shifting loads, there $\exists\delta_{d}$ constants such that $\sum_t  P_{d}=\delta_{d}$, for all $t=[1:T] \in \mathbb{N}$.
\end{assumption}
\begin{observation}In this paper, $P^+$ (e.g. PV generation) is considered a non-curtailable resource.
\end{observation}
\begin{observation}All electrical vehicles, $d_{EV}$, and their associated consumption $P^{-}_{d}$, were considered as time scaling and shifting loads working under the conditions imposed by Assumption~\ref{as:1} and~\ref{as:2}.
\end{observation}

\section{Background and Preliminaries}
In this section, we provide a brief overview of reinforcement learning, Markov decision formalism, and deep neural networks.
\subsection{Reinforcement Learning}
In a Reinforcement Learning (RL)~\cite{Sutton:1998} context, an agent learns to act using a (Partial Observable) Markov Decision Process (MDP) formalism. MDPs are defined by a $4$-tuple $\langle\mathcal{S}, \mathcal{A},\mathcal{T}_\cdot(\cdot,\cdot),$ $\mathcal{R}_\cdot(\cdot,\cdot)\rangle$, where:
\begin{itemize}
\item $\mathcal{S}$ is the state space, $\forall s\in \mathcal{S}$,
\item $\mathcal{A}$ is the action space, $\forall a\in \mathcal{A}$,
\item $\mathcal{T}:\mathcal{S}\times \mathcal{A}\times \mathcal{S}\to [0,1]$ is the transition function given by the probability that by choosing action $a$ in state $s$ at time $t$, the system will arrive at state $s'$ at time $ t+1$, such that $p_a(s,s') = p(s_{t+1}=s' | s_t = s, a_t=a)$, and
\item $\mathcal{R}:\mathcal{S}\times \mathcal{A}\times \mathcal{S}\to \mathbb{R}$ is the reward function, were $\mathcal{R}_a(s,s')$ is the immediate reward received by the agent after it performs the transition to state $s'$ from state $s$.
\end{itemize}
The agent aims to optimize a stochastic policy $\pi:\mathcal{S}\times \mathcal{A}\times \mathcal{R}\to\mathbb{R}_+$.
Under structure assumption of the environment (i.e. finite states and actions) the Markov decision problem is typically solved using dynamic programing. However, in our built environment, the model has a large (continuous) states space. Therein, the state space is given by the building energy consumption and price at every moment in time, while the action space is highly dependent on the \textit{electric device constrains}. The success of every action $a$ is measured by a reward $r$. Learning to act in an environment will make the agent to choose actions to maximize future rewards. The value function $Q^{\pi}(s,a)$ is an expected total reward in state $s$ using action $a$ under a policy $\pi$. Currently, one of the most popular reinforcement learning algorithm is Q-learning~\cite{Watkins}.
\subsection{Deep Neural Networks}
\begin{figure}[h!]  
\centering
\includegraphics[width=0.86\columnwidth]{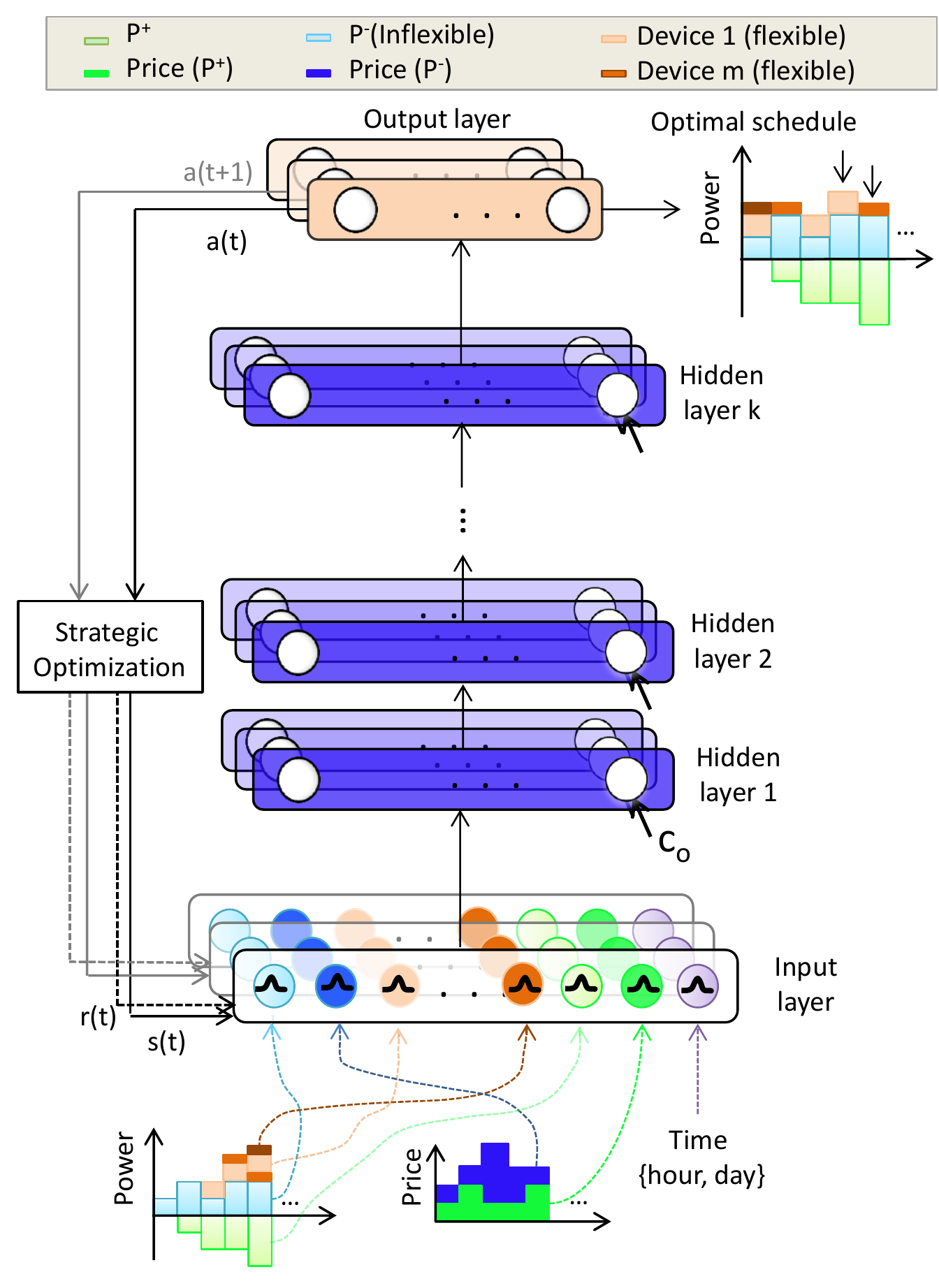}
\caption{The closed loop general architecture of Deep Reinforcement Learning, built as a combination of Reinforcement Learning and Deep Neural Network.}
\label{fig:DRL}
\end{figure}
The topology of a Deep Neural Network (DNN) architecture is based on multiple layers of neurons. In general, a neuron is a non-linear transformation of the linear sum of its inputs. The first layer models directly the data. A  hidden layer in the neural network architecture is build as an array of neurons taking the inputs from the previous layer. The activation function of a neuron on top of $k$ stacked layers in the architecture is using  composite functions, such as $x \otimes h_1\otimes h_2\otimes \dots \otimes h_k$. 

In 2011, it was shown that supervised training of a very deep neural network with hard non-linearities is faster if the hidden layers are composed of Rectified Linear Units (ReLU)~\cite{AISTATS2011_GlorotBB11}. Recently, the logistic sigmoid and the hyperbolic tangent activation are outperformed by ReLU~\cite{relu1,reluHinton,mnih15}. Formally, ReLU is defined as a function $f(x_i)=max(0,x_i)$, where $x_i$ is its input. However, to avoid a non-zero gradient when the hidden units are not active, we used a slightly relaxed form proposed in \cite{7410480}, given by
\begin{equation}\label{eq:relu1}
f(x_i) = \begin{cases}
x_i &\text{if $x_i > 0,\forall i\in \mathbb{N}$}\\
\eta x_i &\text{if $ x_i \le 0, \forall i\in \mathbb{N}$}
\end{cases}
\end{equation}
where $\eta$ is a coefficient controlling the slope of the negative part. If $\eta=0$ then  Eq.~\ref{eq:relu1} becames ReLU. One special case of using a DNN is in deep reinforcement learning where the input is given by the states of an MDP and the output represents the actions of the MDP. 

\section{Proposed Method}
In this section, we propose the use of Deep Reinforcement Learning (DRL) as an on-line method to perform optimal building resource allocation at different levels of aggregation. The general architecture of our proposed method is depicted in Fig.~\ref{fig:DRL}. DRL (RL combined with DNNs of $k$ hidden layers) can learn to act better than the standard RL by automatically extracting patterns, such as those of electricity consumption. Overall, we can represent the DNN method, from a very general perspective, as a black box model with good generalization capabilities over a given input distribution as follows:
\begin{equation}
\xrightarrow[data]{Input} DNN_{(k)} \xrightarrow[Data ~estimation]{Output}
\end{equation}
In the remaining of this section we will introduce two DRL methods, namely Deep Q-learning (DQN) and Deep Policy Gradient (DPG).\\
$DRL= \begin{cases}
\xrightarrow[states]{Input} DNN_{(k)} \xrightarrow[Q(s,a)]{Output}~~\text{Deep Q-learning} \\\nonumber
\xrightarrow[states]{Input} DNN_{(k)} \xrightarrow[p(a|s)]{Output}~~\text{Deep Policy Gradient} 
                    \end{cases}$
In contrast to value-based methods (e.g. DQN), policy-based model free methods (e.g. DPG) directly parameterize the policy $\pi(a|s;\theta)$ and update the parameters $\theta$ by performing, typically approximate, gradient ascent on the expected long-term reward~\cite{Mnih16}. 
 
\subsection{Deep Q-learning (DQN)}\label{Sec:dql} 
Learning in DRL is done as follows. The DNN is trained with a variant of the Q-learning algorithm, using stochastic gradient descent to update its parameters~\cite{mnih15}. Firstly, the value-function from the standard RL algorithm is replaced by a deep Q-network with parameters $\theta$, given by the weights and biases of DNN, such that
$Q(s,a,\theta) \approx Q^{\pi}(s,a)$. This approximation is used further to define the objective function by mean-squared error in Q-values
\begin{equation}\label{eq:1}
\mathcal{L}( \theta)=\mathbb{E}\big[\big(r+\gamma \max_{a_{t+1}}Q(s_{t+1},a_{t+1},\theta)-Q(s_t,a_t,\theta)\big)^2\big]
\end{equation}
Leading to the following Q-learning gradient
\begin{align}\label{eq:q}
\frac{\partial \mathcal{L}(\theta)}{\partial \theta}=\mathbb{E}\Big[\big(r&+\gamma \max_{a_{t+1}}Q(s_{t+1},a_{t+1},\theta)\\ \nonumber&-Q(s_t,a_t,\theta)\big)\frac{\partial Q(s_t,a_t,\theta)}{\partial \theta}\Big]
\end{align}
Usually, this standard Q-learning algorithm used in synergy with neural networks oscillates or diverges, mainly because data are sequential. 
To overcome this limitation of correlated data and non-stationary distributions, we use an \textit{experience replay} mechanism which randomly samples previous mini-batch of transitions $(s_t,a_t,r_t,s_{t+1})$ from the dataset $\mathcal{D}$, and therefore smooths the training distribution over many historical data. It is straightforward to integrate the above Deep RL approach into Eq.1. 
The binary action vector $a_t \in\mathcal{A}$ is augmented to $\max_{a_{t+1}}Q(s_{t+1},a_{t+1},\theta)$ and therefore $\sum_{d=1}^{m}a_{t} P^{-}_{d,t}$ is optimally controlled.  Specifically, rather than enforcing the constraints on the time window required by a specific device $d$ and the comfort of end-users considered in $A1$, Eq.8, our idea is to encapsulate them in the reward function, $r_t(\lambda_{t}^{+},\lambda_{t}^{-}, P^{-}_{i,d})$, which is further detailed in Section IV.A.

\subsection{Deep Policy Gradient (DPG)}
Recently, it has been shown  that policy gradient methods are able to decrease the time needed for convergence in continuous games~\cite{Schulman15, Mnih16}. From an architectural perspective, the neurons from the output layer of the DNN (with parameters $\theta$) corresponding to DPG, instead of estimating the $Q(s_t,a,\theta)$, $\forall a\in\mathcal{A}$ as in DQN, they estimate the probability to take action $a$ in a specific state $s_t$, such as $p(a|s_t,\theta)$, $\forall a\in\mathcal{A}$.
 This offers a clear advantage to DPG over DQN when there is a need to perform multiple actions simultaneously, as all actions can be sampled and executed simultaneously in the game using their own probability. 
 
In the policy gradient context, the approximate optimization problem defined in Eq.~\ref{eq:cost} or Eq.~\ref{eq:12} is an equivalent of maximizing the total expected reward of a parameterized model under a policy $\pi$, as follows  
\begin{equation}\label{eq:122}
\begin{aligned}
& \underset{}{\text{maximize}}
& &  \mathbb{E}_{x\sim p(x|\theta)} [\mathcal{R}|\pi]\\
\end{aligned}
\end{equation}
In the DPG context, the parameterized model is the DNN. Thus, the DNN becomes a probability density function over its inputs (the game states), i.e. $f(x)$, leading Eq.~\ref{eq:122} to the following optimization problem
\begin{equation}\label{eq:122dpg}
\begin{aligned}
& \underset{}{\text{maximize}}
& &  \mathbb{E}_{x\sim p(x|\theta)} [f(x)]\\
\end{aligned}
\end{equation}
As shown in~\cite{Peters2008682, bouramamr2014}, the unbiased gradient estimation uses $f(x)$ as a score function yielding 
\begin{align}\label{eq:gradient}\nonumber
\bigtriangledown_{\theta}\mathbb{E}_{x}[f(x)]&=\bigtriangledown_{\theta}\int\text{dx}~p(x|\theta) f(x)=\int dx \bigtriangledown_{\theta} p(x|\theta) f(x)\\\nonumber
&=\int\text{dx}~p(x|\theta) \frac{\bigtriangledown_{\theta} p(x|\theta)}{p(x|\theta)} f(x)\\\nonumber
&=\int\text{dx}~p(x|\theta)\bigtriangledown_{\theta} \text{log}~p(x|\theta)f(x)\\
&=\mathbb{E}_{x}[f(x)\bigtriangledown_{\theta} \text{log}~p(x|\theta)]
\end{align}
where $\frac{\partial}{\partial \theta}=\bigtriangledown_{\theta}$ denote the first-order partial derivative over the output data.
Intuitively, to solve the gradient of Eq.~\ref{eq:gradient}, first we have to take samples of $x_i\sim p(x|\theta)$ and to compute the estimated gradient, such that $\hat{g}^{\theta}_i=f(x_i)\bigtriangledown_{\theta} \text{log}~p(x_i|\theta)$. Moving in the $\hat{g}_i$ direction increases the log-probability of that particular sample $x_i$ proportional with the reward associated with it, $f(x_i)$. In other words, this practically shows how good is that sample. As in policy gradient the reward is available at the end of a game, these samples are collected in a trajectory, i.e. $\tau=(s_0,a_0,r_0,\dots,s_{T-1},a_{T-1},r_{T-1})$. To compute the gradient of a trajectory, we need to calculate and differentiate the density $p(\tau|\theta)$ with respect to $\theta$ as follows:
\begin{equation}
\label{eq:trajprob}
p(\tau|\theta)=p(s_0)\prod_{t=0}^{T-1}[\pi(a_t|s_t,\theta)p(s_{t+1}|s_t,a_t)]
\end{equation}
By taking the log-probability of Eq.~(\ref{eq:trajprob}), we obtain
\begin{align}\label{eq:log}
\text{log}~p(\tau|\theta)&=\text{log}~p(s_0)+\sum_{t=0}^{T-1}\Big[\text{log}~\pi(a_t|s_t,\theta)\\ \nonumber&~~~~~~~~~~~~~~~+\text{log}~p(s_{t+1}|s_t,a_t)\Big]
\end{align}
Taking the derivative of Eq.~(\ref{eq:log})  with respect to $\theta$ leads to
\begin{equation}
\frac{\partial}{\partial \theta}\text{log}~p(\tau|\theta)=\frac{\partial}{\partial \theta}\sum_{t=0}^{T-1}\text{log}~\pi(a_t|s_t,\theta)
\end{equation}
Finally, we can write the gradient update $\hat{g}^\theta_\tau$ for parameters $\theta$ after considering a trajectory $\tau$ as
\begin{equation}
\hat{g}^\theta_\tau\propto \mathcal{R}_\tau\frac{\partial}{\partial \theta}\sum_{t=0}^{T-1}\text{log}~\pi(a_t|s_t,\theta)
\label{eq:gradientdpg}
\end{equation}

\section{Implementation details}
\subsection{Network Architecture}
Aiming to have a fair comparison between DQN and DPG, the architecture of the deep neural networks used is similar for both models and it has the following characteristics. Each reinforcement learning state (encoded in the input layer), is given by a time-window of two consecutive time steps. Thus, in the case of the peak reduction problem the input layer has 11 neurons, i.e. time step $t$, and base load, PV, AC state, EV and dishwasher at $t-1$ and $t$.
Please note that with the exception of base load and generation which are fixed, the other state components are given by the dynamically adapted values (the ones obtained during the learning process) and not the initial ones measured by the smart meters. For the cost reduction problem, the input layer has an extra neuron which is used to encode the ToU tariff. Furthermore, the networks have three layers of hidden neurons, each layer having 100 neurons with Rectifier Linear Units (ReLU) as activation function. 

The output layer differs for DQN and DPG. For DQN the output layer has 8 neurons, each neuron representing the Q-value of a combined action. Each combined action is a possible combination of the actions of the three flexible devices\footnote{The number of neurons in the output layer of DQN is exponentially correlated with respect to the number of flexible devices.}, i.e. stop air conditioner ($a_1$), electric vehicle on/off ($a_2$), dishwasher on/off ($a_3$). By contrast, the DPG output layer has just three neurons, each neuron representing a device action. More precisely, it gives the probability to perform the action associated with the flexible device for the specific input state. This is a clear advantage of DPG over DQN as it scales linearly with the number of flexible devices.

\textit{Hyper-parameters settings:} In all experiments performed, the learning rate is set to $\alpha=10^{-2}$, the discount factor to $\gamma=0.99$, and  $\eta=0.01$.  We train the models for 5000 episodes, where an episode is composed by 20 randomly chosen days. The weights update is performed after every two episodes. The final policy is kept as the output of the learning process. 

\subsection{The reward vectors for DRL} 
 Regarding the multi-objective optimization problems solved in this paper, an accurate reward function is computed at the end of the day, instead of at each time step of the day. Thus, we derived a simple multiple-task joint reward with three reward components:\\
\textbf{Component 1:} For all $\tau=(s_t,a_t,r_t)$ the reward vectors will be able to control the actions of the three types of flexible consumption, and therefore the total shiftable and scalable load in a household $\sum_{d=1}^{m}a_{t} P^{-}_{d,t}$, using differentiated rewards
\begin{equation}\label{eq:r1}
r_{a_1}=\begin{cases}
-n_{a_1^+} \text{if $n_{a_1^+}>10$ } \\
~~\zeta_1 ~~\text{if $n_{a_1^+}\in[1,10]$;} \\
~~\zeta_2  ~~\text{if $n_{a_1^+}<1$ } 
\end{cases}
r_{a_3} = \begin{cases}
-n_{a_3^+} \text{if $n_{a_3^+}>2$ } \\
~~\zeta_1 ~\text{if $n_{a_3^+}\in[1,2]$ } \\
~~\zeta_2 ~\text{if $n_{a_3^+}<1$ } \\
\end{cases}\nonumber
\end{equation}
\begin{align}\label{eq:r2}
r_{a_2} = \begin{cases}
-4|n_{a_2^t}-n_{a_2^+}| &\text{if $n_{a_2^+}\neq n_{a_2^t}$, $\forall n_{a_2^t}\in\mathbb{N}$ } \\
~n_{a_2^+} &\text{if $n_{a_2^+}=n_{a_2^t}$, $\forall n_{a_2^t}\in\mathbb{N}$ } \\
\end{cases}
\end{align}
where $n_{a_1^+}$, $n_{a_2^+}$, and $n_{a_3^+}$ represent how many times the action corresponding with the flexible device is performed, and $n_{a_2^t}$ is the targeted number of loads per day for the electric car.  The choice of $\zeta_1$ and $\zeta_2$ coefficients  was based on a trial and error procedure. The obtained values are $\zeta_1=40$ and $\zeta_2=-50$.\\
\textbf{Component 2:} Controlling the total energy consumption defined in Eq.~\ref{eq:12} is done as follows
\begin{equation}\label{eq:r40}
\resizebox{.52\textwidth}{!} 
{
$r = \begin{cases}
-3\zeta_2+4[max(P^{-})-max(\tilde{P}^{-})]~\text{if $max(\tilde{P}^{-})<max(P^{-})$ } \\
-3\zeta_1-1~~~~~~~~~~~~~~~~~~~~~~~~~~~~~~~~\text{otherwise} 
\end{cases}$
}
\end{equation}
Further on, shifting the consumption through the time when there is more generation Eq.~\ref{eq:const}.
\begin{equation}\label{eq:r4}
r = \begin{cases}
~~\frac{\zeta_1}{2}-|min(\tilde{P}^{-})|~~\text{if $\tilde{P}^{-}<0$ } \\
-\frac{\zeta_2}{2}~~~~~~~~~~~~~~~~~~~\text{otherwise} 
\end{cases}
\end{equation}
The control of AC under the A\ref{as:2}, Eq.~\ref{eq:ass1} is given by
\begin{align}\label{eq:r5}
r= \begin{cases}
\frac{\zeta_1}{8}+2[max(\tilde{P}_{AC}^{-})-max(P_{AC}^{-})]&\text{if $\tilde{P}^{-}<0$ } \\
-\frac{\zeta_2}{10}&\text{otherwise} 
\end{cases}
\end{align}
\textbf{Component 3:} Controlling the total cost $\mathcal{C}$, defined in Eq.~\ref{eq:cost}.
\begin{align}\label{eq:r6}
 r = \begin{cases}
5|\tilde{\mathcal{C}}-\mathcal{C}|&\text{if $\tilde{\mathcal{C}}<\mathcal{C}$ } \\
-3\zeta_1-1&\text{otherwise} 
\end{cases}
\end{align}

The agent must learn multiple tasks consecutively with the goal of optimizing performance across all previously learned tasks.
So, we used for solving  Eq.~\ref{eq:12} the \textbf{Component 1} and \textbf{2} of the reward, while for Eq.~\ref{eq:cost} the \textbf{Component 1} and \textbf{3}.  

The joint reward components could be easily generalized to perform an arbitrary number of tasks. However, the range intervals for $n_{a_1^+}$ and $n_{a_3^+}$ considered in Eq.~\ref{eq:r1} as well as the \textit{positive} and \textit{negative} coefficients (i.e. $\zeta_1$ and $\zeta_2$) used in Eq.~\ref{eq:r1}-\ref{eq:r6} are dependent on the application. Also in Eq.~\ref{eq:r1} the range of $n_{a_1^+}$ and $n_{a_3^+}$ may be enlarged if comfort limits are relaxed. Algorithm~\ref{algo:dpg} exemplifies on DPG, how DRL can be implemented. We have implemented both methods, DQN and DPG, in Python. 
\begin{algorithm}
\small
 \caption{Deep Policy Gradient (DPG) - estimating Eq.\ref{eq:122}}
 \begin{algorithmic}[1]
 \renewcommand{\algorithmicrequire}{initialize: DNN with random weights $\theta$ and Q}
\STATE initialize model: hyper-parameters ($\alpha$, $\gamma$, $\zeta$)
\STATE initialize model: DNN with random weights $\theta$
\STATE initialize game: first state $s$ from a random day 
  \FOR { $iteration=1$ to $arbitrary$ $number$}
  \STATE sample actions $p(a_1,a_2,a_3|\theta,s)$ with DNN
  \STATE collect probabilities  $p(a_1,a_2,a_3|\theta,s)$ in $A$
  \STATE collect hidden neurons values in $H$ from DNN
  \STATE collect $s$ in $S$ 
  \STATE execute actions $a_1, a_2, a_3$ and move to next state $s'$
  \STATE collect reward $r$ in $R$ from game
  \IF {episode is finished}
  \STATE compute discounted rewards $R_d$ from $R$
  \STATE estimate gradients from $A*R_d$, $\theta$, $S$, and $H$ (Eq.~\ref{eq:gradientdpg})
  \STATE update $\theta$ with the estimated gradient
  \STATE empty $A$, $R$, $S$, and $H$
  \ENDIF
  \IF {current day ends}
  \STATE reset game: first state $s'$ from a random day 
  \ENDIF
  \STATE set $s=s'$
  \ENDFOR
 \end{algorithmic} 
 \label{algo:dpg}
 \end{algorithm}

\section{Results and Discussion}
In this section, we validate our proposed methods and we analyze their performance on a large real-world database recorded by Pecan Street.Inc. First, the database is described. Then, numerical results are given for both problems, i.e. peak reduction and cost minimization for various number of buildings. 
\subsection{Data set characteristics}
\subsubsection{Buildings pattern}
\begin{figure*}
\centering
\subfloat[Electrical patterns~\textendash~Building I]{\includegraphics[width=2.39in]{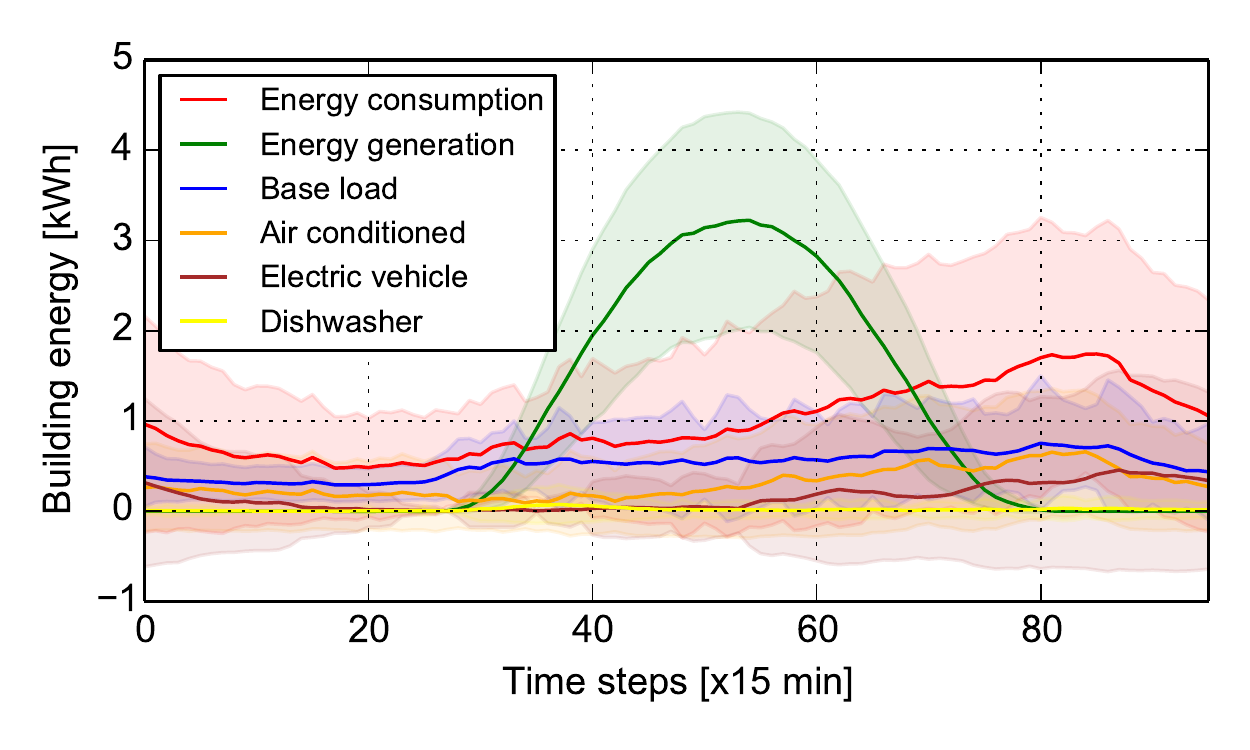}%
\label{fig_first_case}}
\subfloat[Electrical patterns~\textendash~Building II]{\includegraphics[width=2.39in]{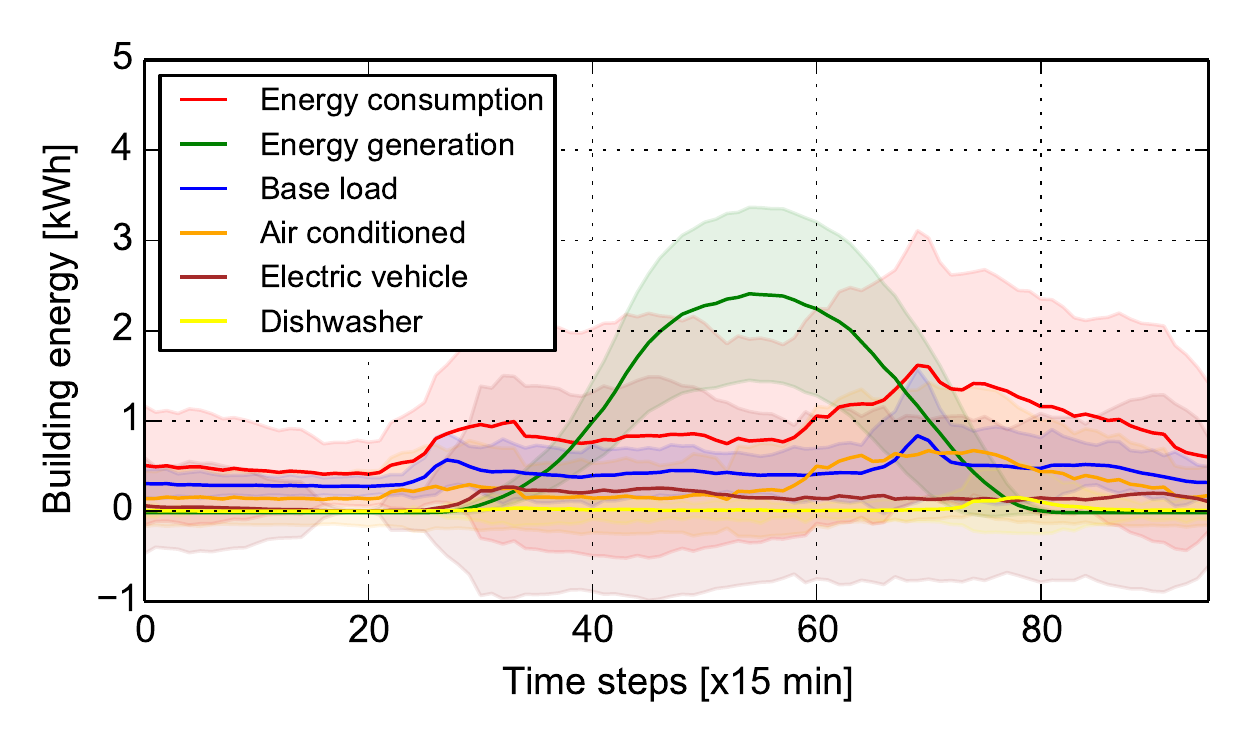}%
\label{fig_second_case}}
\subfloat[Electrical patterns~\textendash~Building III]{\includegraphics[width=2.39in]{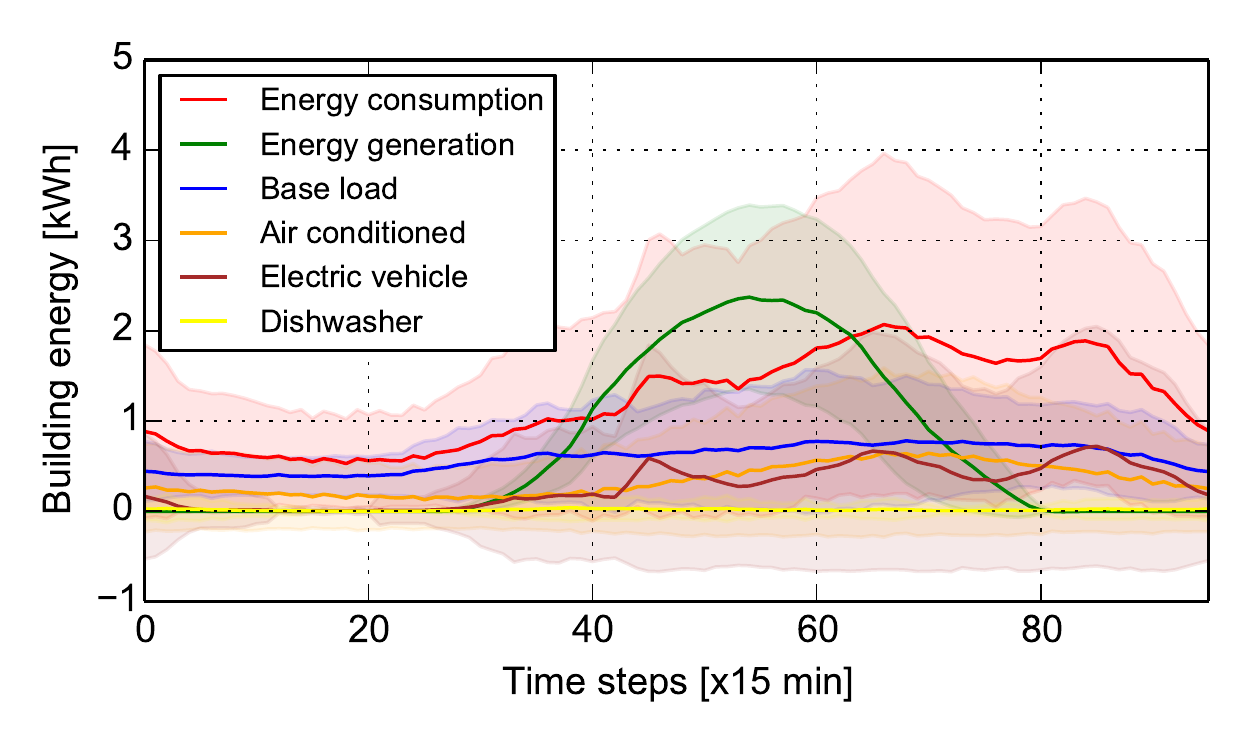}%
\label{fig_3_case}}\\
\subfloat[Peak reduction using DPG~\textendash~Building I]{\includegraphics[width=2.39in]{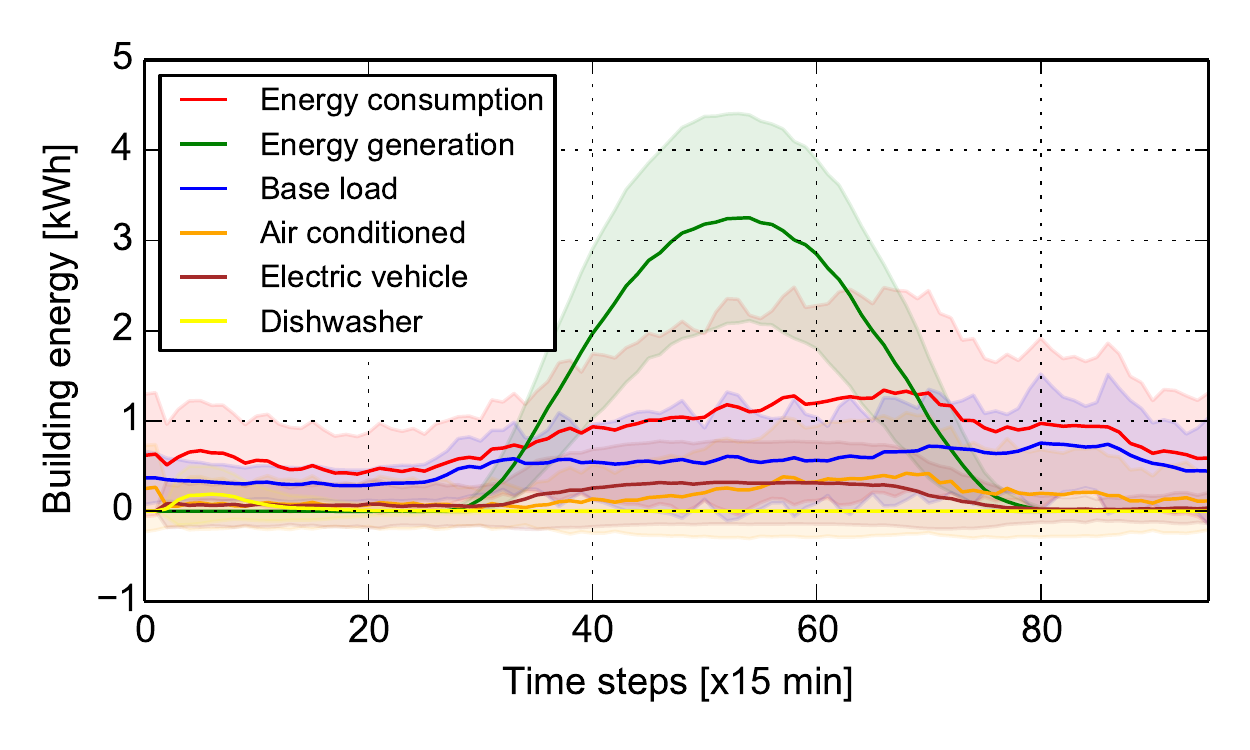}%
\label{optimized_fig_first_case}}
\subfloat[Peak reduction using DPG~\textendash~Building II]{\includegraphics[width=2.39in]{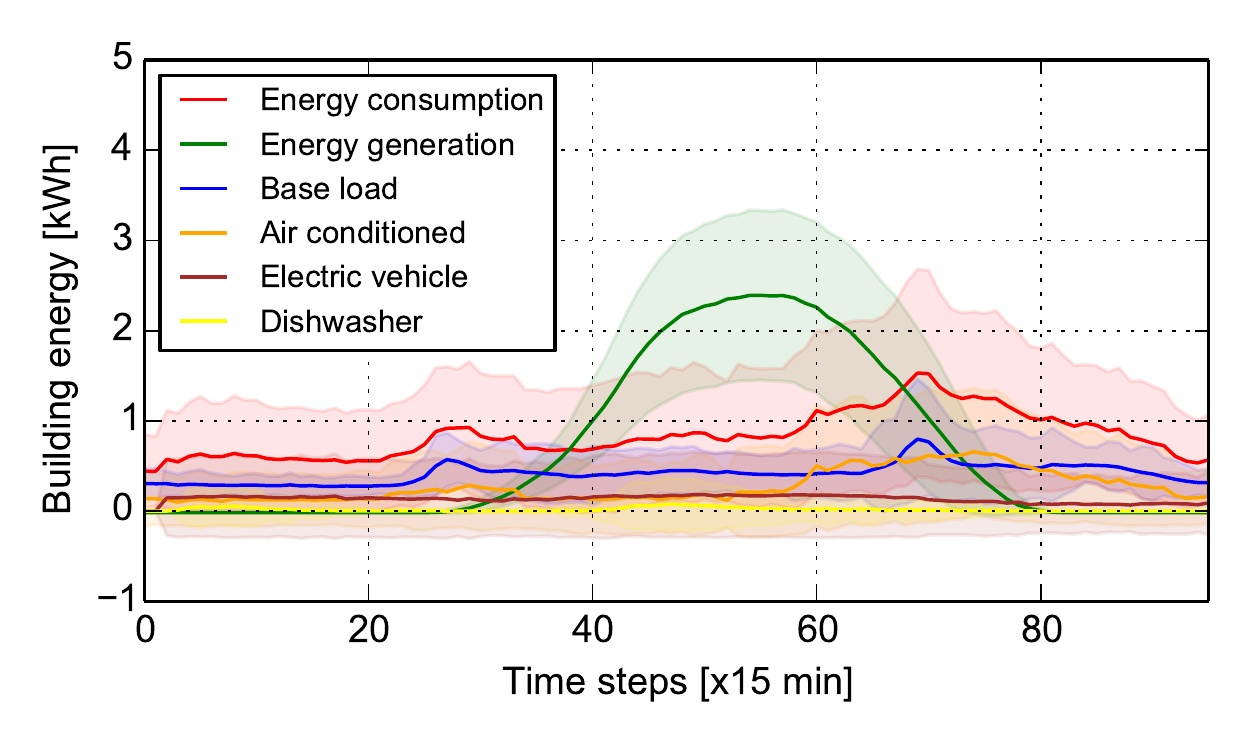}%
\label{optimized_fig_second_case}}
\subfloat[Peak reduction using DPG~\textendash~Building III]{\includegraphics[width=2.39in]{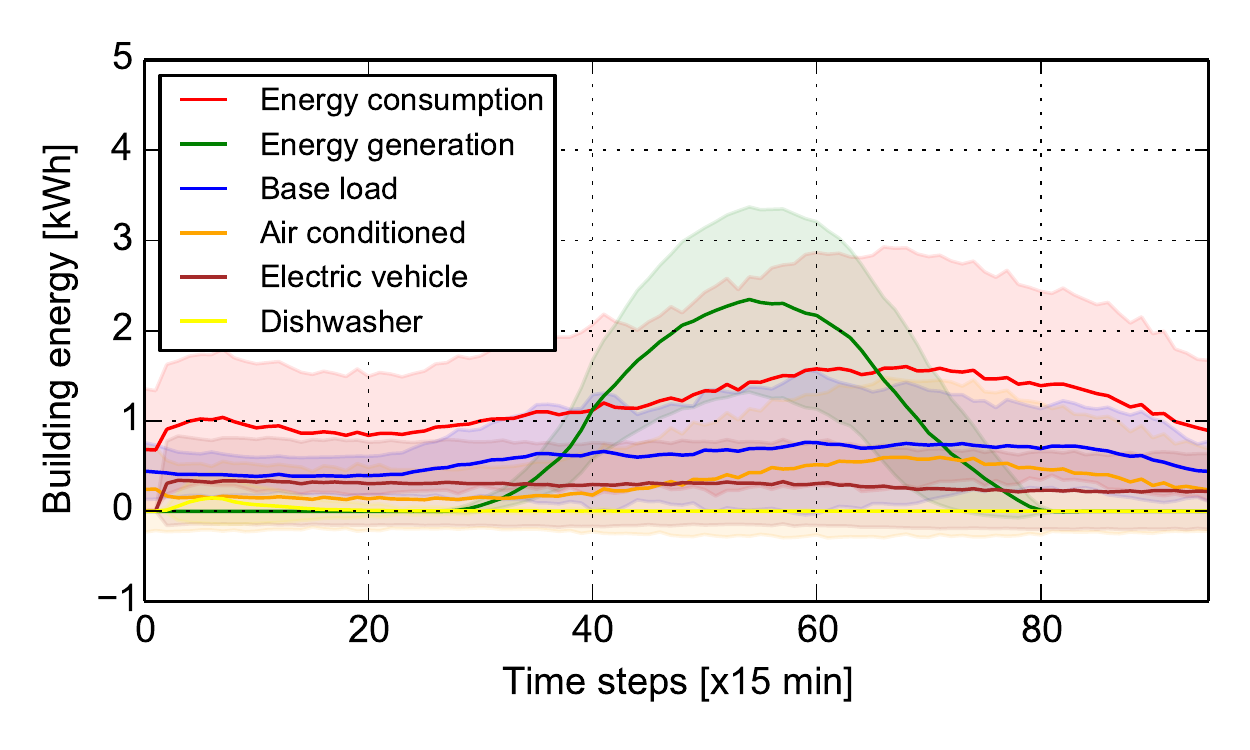}%
\label{optimized_fig_3_case}}\\
\subfloat[Cost minimization using DPG~\textendash~Building I]{\includegraphics[width=2.39in]{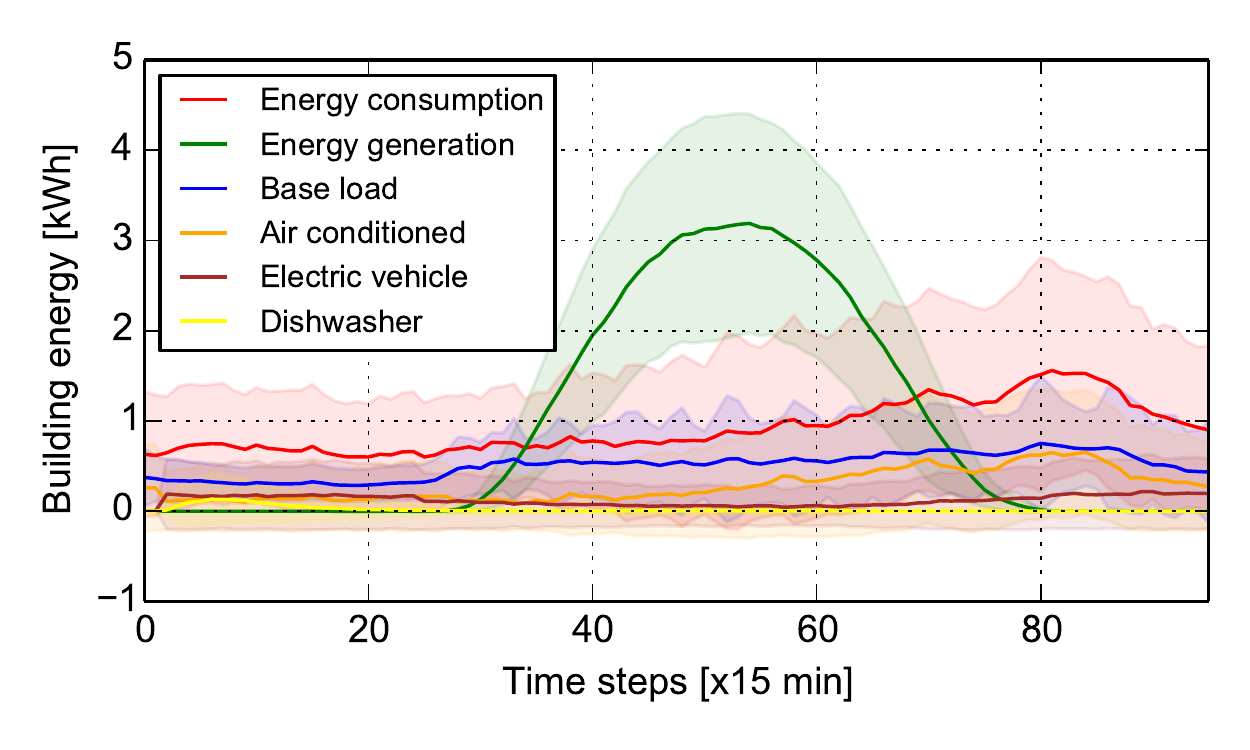}%
\label{cost_optimized_fig_first_case}}
\subfloat[Cost minimization using DPG~\textendash~Building II]{\includegraphics[width=2.39in]{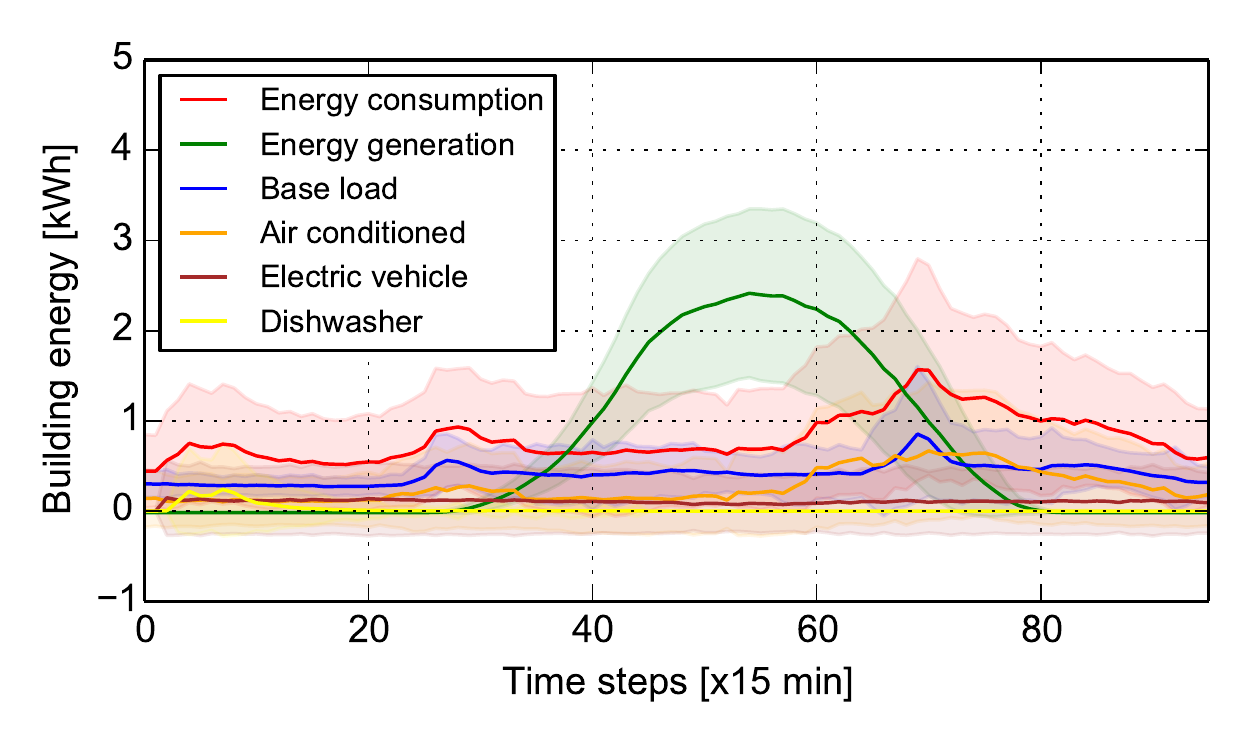}%
\label{cost_optimized_fig_second_case}}
\subfloat[Cost minimization using DPG~\textendash~Building III]{\includegraphics[width=2.39in]{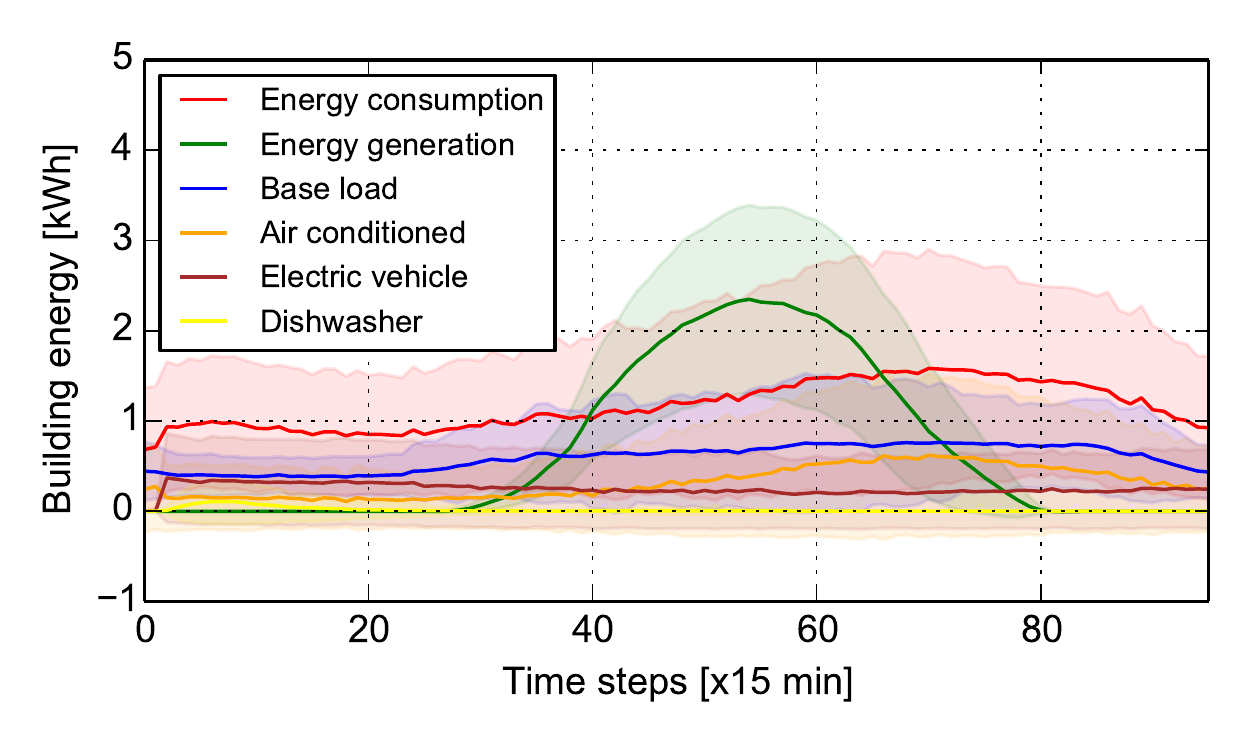}%
\label{cost_optimized_fig_3_case}}
\caption{(a), (b) and (c) represent different building electrical patterns averaged over one year (solid line) with 15 minute resolution, followed by their standard deviation (shadow area). Their yearly average optimization results using Deep Policy Gradient method for the peak reduction problem are depicted in Fig.2 (d), (e) and (f) whether the cost minimization results are showed in Fig.2. (g), (h) and (i).}
\label{fig_buildings}
\vspace*{-2mm}
\end{figure*}
To validate our proposed method we used the Pecan Street dataset. The disaggregated customer energy data contains up to 90 million unique electricity consumption records per day, which are used in order to build specific device patterns. Figure~\ref{fig_buildings}(a), \ref{fig_buildings}(b), and \ref{fig_buildings}(c)  show three different building patterns averaged over the year 2015 with 15 minutes resolution. In these patterns the solar generation uncertainty as well as the people behavior characteristics are notable, e.g. even if all three buildings have an electric vehicle, just in the case of the third building, Fig.~\ref{fig_buildings}(c), it is used frequently. In our experiments, we have used the data between 27th October 2012 and 3rd September 2016.

\subsubsection{Price data} 
We use the time-of-use (ToU) tariff provided by the local grid operator Austin Energy for customers who live inside the City of Austin, Texas\footnote{http://austinenergy.com/wps/portal/ae/residential/rates/residential-electric-rates-and-line-items (Last visit: 30 October 2016)}. The $\lambda^-$ summer  rates are composed by on-peak, mid-peak and off-peak hours and the winter tariff has the mid-peak and off-peak hours components. There is also a difference between weekend and working-days tariff. Additionally, the self-generating customers are receiving an amount that is being paid by the utility for solar generation, called the value of solar tariff (VOST). 
\subsection{Numerical results - Peak reduction problem}
The numerical results in terms of peak reduction at the single building level are showed in Table~\ref{tab:table_year_peak_bullding} and Figure~\ref{fig_buildings} for three different buildings ($B_{I}$, $B_{II}$ and $B_{III}$), over one year with 15 minutes resolution. 
\renewcommand{\tabcolsep}{4pt}
\begin{table}
\caption{Daily peak value at the building level $(B)$ averaged over one year with 15 minutes resolution versus optimized peak value using DQN and DPG methods}
\label{tab:table_year_peak_bullding}
\centering
\begin{tabular}{|c|c|c|c|c|c|c|c|}
\hline
&&\multicolumn{2}{c}{$B_{I}$}&\multicolumn{2}{c}{$B_{II}$}&\multicolumn{2}{c|}{$B_{III}$}\\
\cline{3-8}
&Method&Mean&St.dev.&Mean&St.dev.&Mean&St.dev.\\
&&($\mu$)&($\Sigma$)&($\mu$)&($\Sigma$)&($\mu$)&($\Sigma$)\\
\hline
Peak [kW]&-&3.81&1.72&3.77&2.32&4.55&1.52\\
\hline
Optimized&DQN&2.72&1.45&2.72&1.21&3.59&1.41\\
peak [kW]&DPG&2.55&1.36&2.49&1.13&3.12&1.31\\
\hline
\end{tabular}
\vspace*{-1mm}
\end{table}
 
\subsection{Numerical results - Cost minimization problem}
The results for the cost minimization problem are summarized in Table~\ref{tab:table_year_cost_building}. The difference between the cost minimization solutions obtained for every building correlated with their average electrical patterns (see Figure~\ref{fig_buildings}) give as a first indication about the individual capabilities of the end-users to adopt a more conservative behavior. 

As it can be observed in Table~\ref{tab:table_year_cost_building} and in Figure~\ref{fig_buildings}, a secondary advantage of solving the cost minimization problem is its impact on solving of the peak reduction problem also. Therein, the best results in terms of both, peak reduction and cost reduction, are obtained for building $B_{III}$ using DPG.  
\renewcommand{\tabcolsep}{3.5pt}
\begin{table}
\caption{Daily cost minimization results at the building level average over one year with 15 minutes resolution using DQN and DPG methods}
\label{tab:table_year_cost_building}
\centering
\begin{tabular}{|c|c|c|c|c|c|c|c|}
\hline
&&\multicolumn{2}{c}{$B_{I}$}&\multicolumn{2}{c}{$B_{II}$}&\multicolumn{2}{c|}{$B_{III}$}\\
\cline{3-8}
&Method&Mean&St.dev.&Mean&St.dev.&Mean&St.dev.\\
&&($\mu$)&($\Sigma$)&($\mu$)&($\Sigma$)&($\mu$)&($\Sigma$)\\
\hline
Peak [kW]&-&3.81&1.72&3.77&2.32&4.55&1.52\\
\hline
Peak [kW] &DQN&3.12&1.51&3.48&2.13&3.62&1.34\\
reduction&DPG&2.97&1.46&2.69&1.14&3.17&1.29\\
\hline
\hline
Cost [\$/day]&-&2.31&3.09&1.93&2.23&3.13&3.85\\
\hline
Minimized&DQN&2.19&3.01&1.91&2.18&2.85&3.62\\
cost [\$/day]&DPG&2.08&2.78&1.79&2.06&2.73&3.38\\
\hline
\end{tabular}
\end{table}
\subsection{Scalability and learning capabilities of DRL}
\begin{figure}
\centering
\includegraphics[width=3.4in]{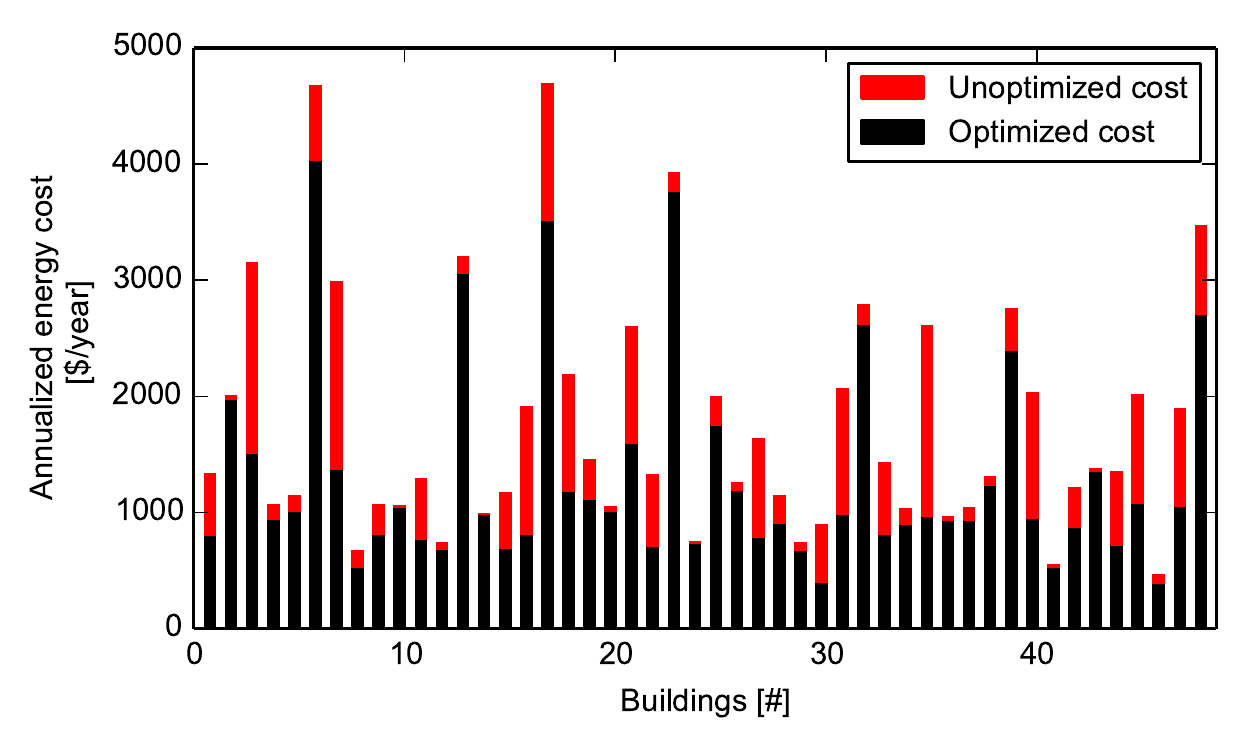}
\vspace*{-4mm}
\caption{Yearly savings per buildings when cost optimization is performed at the aggregated level on 48 buildings using Deep Policy Gradient (DPG) method.}
\label{fig_totalcost}
\vspace*{-6mm}
\end{figure}

To test whether good estimations occur in practice and at scale, we investigate the performance of our proposed methods, in three cases with different numbers of customers using data from the Pecan Street smart grid test-bed. Specifically, we are  investigating and analyzing the corresponding results using DQN and DPG methods for 10, 20 and 48 buildings, respectively. Tables~\ref{table_year_peak} and~\ref{table_year_cost} show that our proposed approach is scalable for both, peak reduction and cost minimization, respectively. 
More than that, they show that at the aggregated level, when more customers are taking the cost minimization problem into consideration, this solves implicitly also the peak reduction problem.  
Same as at the building level, DPG is more stable than DQN, and achieves a better performance. Overall, in the case of the cost reduction problem for 48 buildings, DPG reduces the peak with 26.3\% and minimizes the cost with  27.4\%, while DQN reduces the peak with just 9.6\% and minimizes the cost with just 14.1\%. To visualize how DPG performs, in Figure~\ref{fig_totalcost} we depict the unoptimized and the optimized annualized energy costs for each of the 48 buildings. 
\renewcommand{\tabcolsep}{3.5pt}
\begin{table}
\caption{Peak reduction $-$ Daily optimization results at different levels of aggregation average over one year with 15 minutes resolution using DQN and DPG methods}
\label{table_year_peak}
\centering
\begin{tabular}{|c|c|c|c|c|c|c|c|}
\hline
&&\multicolumn{6}{c|}{Number of buildings}\\
&&\multicolumn{2}{c}{10}&\multicolumn{2}{c}{20}&\multicolumn{2}{c|}{48}\\
\cline{3-8}
&Method&Mean&St.dev.&Mean&St.dev.&Mean&St.dev.\\
&&($\mu$)&($\Sigma$)&($\mu$)&($\Sigma$)&($\mu$)&($\Sigma$)\\
\hline
Peak [kW]&-&59.79&6.12&124.72&10.28&281.88&14.32\\
\hline
Optimized&DQN&49.67&5.62&106.84&7.49&238.12&12.98\\
peak [kW]&DPG&41.74&5.08&93.83&7.29&213.01&12.02\\
\hline
\end{tabular}
\end{table}
\renewcommand{\tabcolsep}{3.5pt}
\begin{table}
\caption{Cost reduction $-$ Daily optimization results at different levels of aggregation average over one year with 15 minutes resolution using DQN and DPG methods}
\label{table_year_cost}
\centering
\begin{tabular}{|c|c|c|c|c|c|c|c|}
\hline
&&\multicolumn{6}{c|}{Number of buildings}\\
&&\multicolumn{2}{c}{10}&\multicolumn{2}{c}{20}&\multicolumn{2}{c|}{48}\\
\cline{3-8}
&Method&Mean&St.dev.&Mean&St.dev.&Mean&St.dev.\\
&&($\mu$)&($\Sigma$)&($\mu$)&($\Sigma$)&($\mu$)&($\Sigma$)\\
\hline
Peak [kW]&-&59.79&6.12&124.72&10.28&281.88&14.32\\
\hline
Peak [kW]&DQN&54.85&5.93&116.72&9.24&254.67&13.21\\
reduction&DPG&44.91&4.80&92.41&7.74&207.73&11.48\\
\hline
\hline
Cost [\$/day] &-&57.79&20.90&118.03&30.01&231.27&38.76\\
\hline
Minimized &DQN&47.71&17.83&93.68&24.18&198.51&32.67\\
cost [\$/day]&DPG&44.35&16.01&82.71&21.48&167.70&28.62\\
\hline
\end{tabular}
\vspace*{-4mm}
\end{table}
To visualize how DPG performs, in Figure~\ref{fig_totalcost} we depict the unoptimized and the optimized annualized energy costs for each of the 48 buildings. We can observe that the buildings behave very differently and in some cases DPG is capable to halve the yearly cost, while in other cases it succeeds to reduce the cost with just a few percentage points.

\begin{figure}
\centering
\includegraphics[width=3.4in]{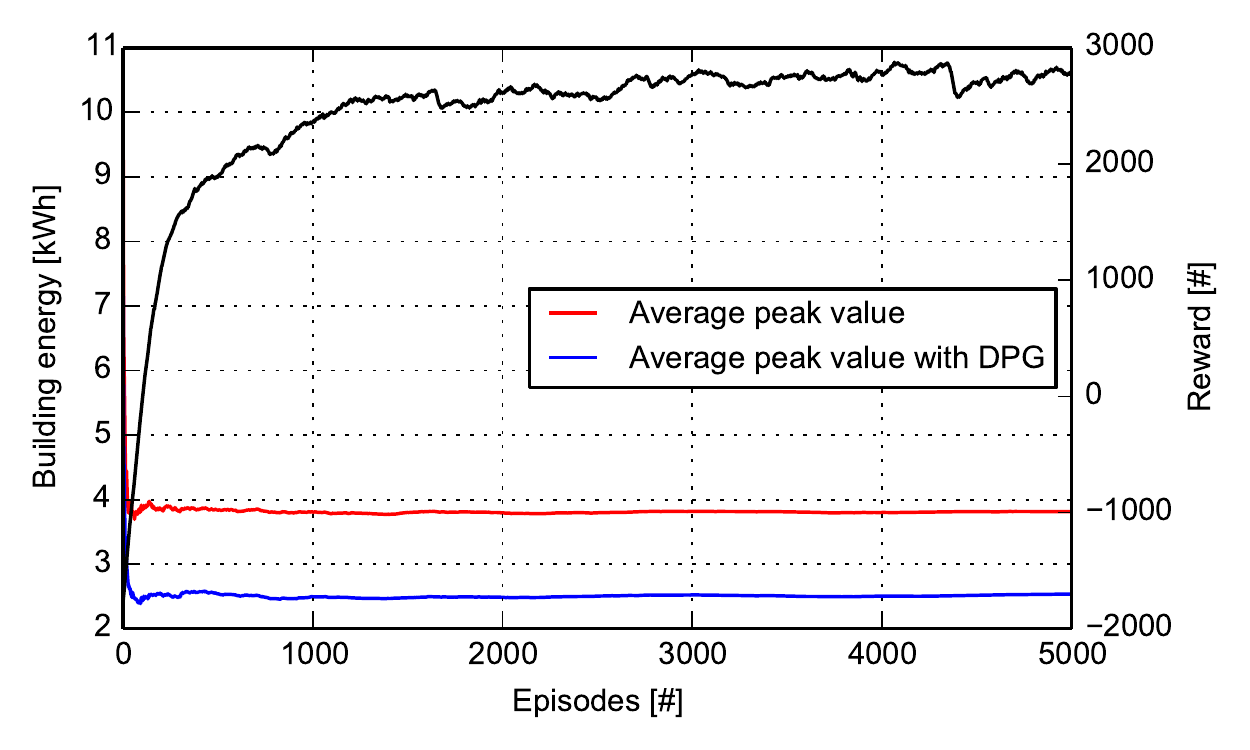}
\vspace*{-4mm}
\caption{Learning capabilities of Deep Policy Gradient method in terms of peak reduction and their corresponding reward function for a building (i.e. Fig.2 Building I). Every epoch represent an average value over 20 random days.}
\label{fig_error}
\vspace*{-4mm}
\end{figure}

\textit{Convergence capabilities of DPG:}
The convergence is assessed through many iterations over episodes. For example, the learning capabilities of DPG method in terms of peak reduction and their corresponding reward function for a building are showed in Figure~\ref{fig_error}.  Each episode represents an average value over 20 randomly chosen days. Initially, we may observe that the reward increases fast, while after about 1000 episodes the reward, as expected, increases much slower. Therefore, after approximatively 1000 episodes the average peak value and the optimized average peak value using the Deep Policy Gradient method converge. Still, the long-term reward expectation, as was expressed in Eq.~(\ref{eq:122}),  is increasing until  approximatively 2500 episodes. 

\textit{Computational time requirements:}  
Both DRL variants have the advantage of handling naturally much larger continuous state spaces, leading to better performance. In comparison with heuristic methods (e.g. PSO), after DRL learns how to act, it can make decisions (e.g. choosing the optimal control action) in a few milliseconds, while PSO needs to re-run the costly optimization process for every decision. 
\section{Conclusions}
In this paper, we proposed the use of Deep Reinforcement Learning, as a hybrid method which combines Reinforcement Learning with Deep Learning, with the aim of conceiving an on-line optimization for the scheduling of electricity consuming devices in residential buildings and aggregations of buildings. We have shown that a single agent, empowered with a suitable learning algorithm, can solve many challenging tasks. We proposed two optimization methods, Deep Q-learning and Deep Policy Gradient, to solve the same sequential decision problems at both the building level and the aggregated level. At both levels, we showed that Deep Policy Gradient is more suited to perform on-line scheduling of energy resources than Deep Q-learning. We explored and validated our proposed methods using the large Pecan Street database. Both methods are able to successfully perform either the minimization of the energy cost or the flattening of the net energy profile. For the minimization of the energy cost, a variable electricity price signal is investigated to incentivize customers to shift their consumption to low-price, off-peak periods.

\section*{Acknowledgment}
This research has been partly funded by the NL Enterprise Agency under the TKI SG-BEMS project of
Dutch Top Sector and by the European Union's Horizon 2020 project INTER-IoT (grant number 687283). 

\ifCLASSOPTIONcaptionsoff
  \newpage
\fi

 \bibliographystyle{IEEEtran}


\end{document}